\documentclass[10pt,twocolumn,letterpaper]{article}

\usepackage{iccv}
\usepackage{times}
\usepackage{graphicx}
\usepackage{amsmath}
\usepackage{amssymb}
\usepackage{algorithm}
\usepackage[algo2e]{algorithm2e}
\usepackage[noend]{algpseudocode}
\usepackage{multirow}
\usepackage{listings}
\usepackage{booktabs}

\usepackage[pagebackref=true,breaklinks=true,colorlinks,bookmarks=false]{hyperref}

\newcommand{\squishlist}{
 \begin{list}{$\bullet$}
  { \setlength{\itemsep}{0pt}
     \setlength{\parsep}{1pt}
     \setlength{\topsep}{1pt}
     \setlength{\partopsep}{0pt}
     \setlength{\leftmargin}{1.2em}
     \setlength{\labelwidth}{1em}
     \setlength{\labelsep}{0.5em} } 
}

\newcommand{\squishend}{
  \end{list}  
}

\iccvfinalcopy

\begin{document}

\title{Instance-based Max-margin for Practical Few-shot Recognition}

\author{Minghao Fu, Ke Zhu and Jianxin Wu \\
State Key Laboratory for Novel Software Technology \\
Nanjing University, Nanjing, China \\
{\texttt{\{fumh,zhuk\}@lamda.nju.edu.cn, wujx2001@gmail.com}}
}

\maketitle

\begin{abstract}
In order to mimic the human few-shot learning (FSL) ability better and to make FSL closer to real-world applications, this paper proposes a practical FSL (pFSL) setting. pFSL is based on unsupervised pretrained models (analogous to human prior knowledge) and recognizes many novel classes simultaneously. Compared to traditional FSL, pFSL is simpler in its formulation, easier to evaluate, more challenging and more practical. To cope with the rarity of training examples, this paper proposes IbM2, an instance-based max-margin method not only for the new pFSL setting, but also works well in traditional FSL scenarios. Based on the Gaussian Annulus Theorem, IbM2 converts random noise applied to the instances into a mechanism to achieve maximum margin in the many-way pFSL (or traditional FSL) recognition task. Experiments with various self-supervised pretraining methods and diverse many- or few-way FSL tasks show that IbM2 almost always leads to improvements compared to its respective baseline methods, and in most cases the improvements are significant. With both the new pFSL setting and novel IbM2 method, this paper shows that practical few-shot learning is both viable and promising. 
\end{abstract}

\section{Introduction}

We human have the ability to learn new concepts based on a few exemplars, thanks to both our ability to learn and the previously accumulated knowledge. As a stark contrast, learning machines (especially deep learning models) mostly require plentiful training examples to learn just few (\eg, 5) concepts. Both the vision and learning community are fully aware of this shortcoming, and few-shot learning (FSL) is our community's effort to counter this weakness, which has been studied for a long time~\cite{wang2020generalizing}. 

FSL relies on prior knowledge, too. Currently, the setting is to set aside a collection of object categories (known as the \emph{base set}), with many training examples in every base set category. A model is first pretrained based on the base set using either self-supervised learning~\cite{su2020does,lu2022self,yang2022few,hu2022pushing,luo2021rectifying}, meta-learning~\cite{lee2019meta,snell2017prototypical,finn2017maml,sung2018learning,vinyals2016matching}, or traditional supervised approaches~\cite{mangla2020charting,dhillon2020baseline,chen2019closer}. The model naturally encodes prior knowledge from the base set. Then, few \emph{novel} concepts or object categories (most commonly 5) are to be recognized based on the pretrained model and few training examples from these new concepts (\eg, 1 or 5 per category). It is worth noting that the base and the novel training sets are semantically closely related (\eg, both contain different kinds of birds). This setting is also \emph{complicated}.

There is a clear mismatch between the human ability and this FSL setting. We human can learn \emph{many} instead of few novel concepts; and, the human prior knowledge include those accumulated as both common-sense and domain knowledge from \emph{many diverse domains} rather than a limited set of  concepts closely related to a specific task.

Although this FSL setting has served our community well and has pushed the technical frontier for years, with the immense advances in deep learning and emerging practical needs for real-world FSL applications, \emph{it is time that we need a better, simpler, and more practical FSL setting.} 

Hence we advocate a practical few-shot learning (or pFSL) setting: based on an \emph{unsupervised} model pretrained from a large number of concepts (\eg, ImageNet~\cite{russakovsky2015imagenet}), simultaneously learn \emph{many} new concepts (\eg, 200) with \emph{few} examples (\eg, 1 or 5 per category). 

An ImageNet pretrained model is more analogous to human knowledge than those learned from a base set in traditional FSL, and learning many new concepts makes it more practical in applications. \emph{Without the base set}, the learning phase of pFSL is simple, too. More importantly, evaluation of algorithms is difficult, complicated, and very time consuming in the current FSL setting~\cite{fu2022acsr}, but as will be discussed later, pFSL does not suffer from this drawback.

Although an ImageNet pretrained model undoubtedly contains more prior knowledge than a base-set trained one, simultaneously learning many new concepts will inevitably make pFSL much more difficult than traditional FSL. We believe that a more challenging task will help technology advancement, too---the traditional FSL setting is already saturated to certain extent (\cf results in~\cite{yang2022few,hu2022pushing}).

Hence, we propose a novel approach for pFSL: Instance-based Max-margin (IbM2), which is effective in traditional FSL, too. The max-margin idea has been shown to be effective in improving generalization. Classic methods like SVM~\cite{chang2011libsvm} build max-margin decision boundaries with \emph{only a very small portion} of the training examples (\ie, support vectors). This idea is problematic when we have both very few training examples and a very high dimensionality. Intuitively, the support vectors chosen by SVM are highly likely wrong or misleading, because a small training set leads to excessively unstable estimation~\cite{fu2022acsr}.

\begin{figure}
	\centering
	\includegraphics[width=0.99\linewidth]{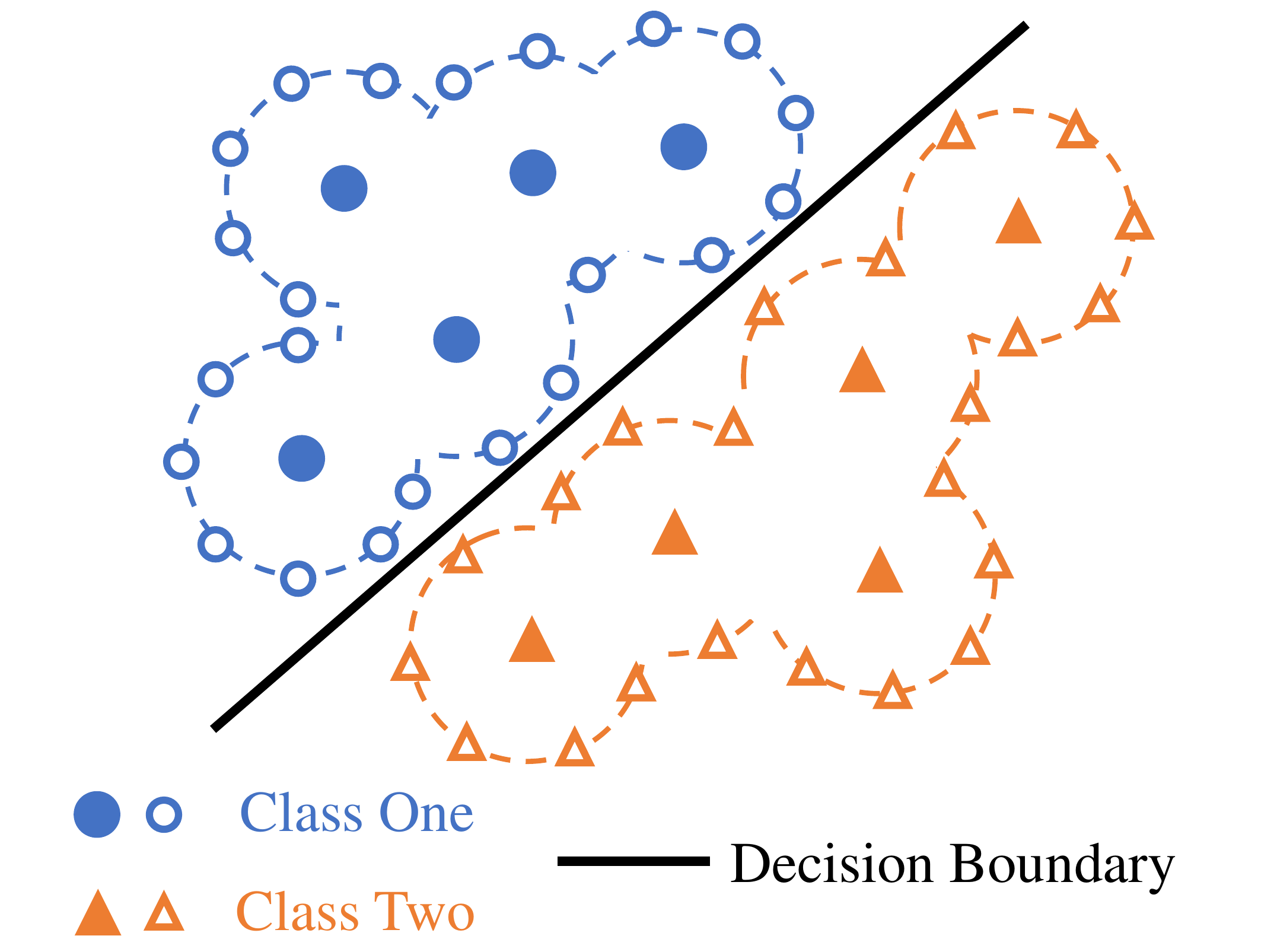}
	\caption{Instance-based max-margin. We create a hypersphere centered at every example (solid shapes), sample many virtual examples (hollow shapes) around the hypersphere, and require the decision boundary to classify all virtual samples correctly. Then we in effect achieve instance-based max-margin. The large dashed circles denote hypersphere. Best viewed in color.}
	\label{fig:intro_sketch}
 \end{figure} 

In the proposed IbM2, we achieve max-margin \emph{at the instance level}, instead of at the class level as in SVM. As illustrated in Fig.~\ref{fig:intro_sketch}, we create a hypersphere for \emph{every} training example, with the example being the center of the hypersphere. Then, we can sample as many \emph{virtual samples} as we want around this hypersphere, and assign the label of the center (original training example) to all its associated virtual examples. By finding a properly \emph{maximized radius} for the hypersphere and requiring a classifier to \emph{correctly classify the virtual examples}, we in effect achieve max-margin. Another benefit of IbM2 is that we do not need any special handling for multi-class recognition, while vanilla SVM is for binary classification only.
In short, the contributions of this paper are twofold:
\squishlist
   \item We propose a practical few-shot learning setting (pFSL): many-way (\eg, 200-way) recognition, uses an unsupervised pretrained model, and has no base set.
   \item We propose IbM2, an instance-based max-margin algorithm, which suits few-shot, high dimensionality, and multi-class naturally.
\squishend
As will be shown by extensive experiments, IbM2 consistently improves both traditional FSL and the proposed pFSL. The benefits of pFSL against traditional FSL will be verified by both analyses and experiments, too.

\section{Related Work}

Few-shot learning aims at recognizing novel categories with the help of some form of prior knowledge. Given the base set as prior knowledge, few-shot learning methods can be roughly divided into two types: those based on meta-learning or transfer-learning.

\textbf{Meta-learning based FSL.} Meta-learning~\cite{hospedales2021meta_survey} learns a model through a large quantity of episodes consisting of different training and evaluation sets. Given a new task during evaluation, the model performs what it does in training to predict. In FSL, the training episodes are randomly sampled from the base set. One line of work~\cite{finn2017maml,lee2019meta,finn2018probabilistic,rusu2019meta} focuses on learning general initial weights for fast adapting the model to unseen categories with a few steps of optimization. For example,~\cite{finn2017maml} explicitly trains the classifier in a model-agnostic way. \cite{lee2019meta} extends meta-learning with linear predictors as a differentiable convex problem. \cite{finn2018probabilistic} learns the task distribution from a probabilistic perspective. Another line of work~\cite{snell2017prototypical,vinyals2016matching,sung2018learning,ye2020few,zhang2022deepemd,afrasiyabi2022matching,chen2021meta} boosts meta-learning by exploring different distance metrics. For example, \cite{snell2017prototypical} aggregates training features to generate prototypes for different classes as the classification head. \cite{zhang2022deepemd} leverages the Earth Mover's Distance to calculate the structural distance for classification. \cite{afrasiyabi2022matching} extracts sets of features from each image to build a set-based matching schema. \cite{chen2021meta} proposes a meta-baseline, which performs pretraining and meta-training sequentially.

\textbf{Transfer-learning based FSL.} This line of attack~\cite{chen2019closer,mangla2020charting,dhillon2020baseline,yang2021free,tian2020rethinking} leverages the idea of standard transfer learning, which first pretrains a model on base classes and then fine-tunes the model weights with limited training samples from novel classes. \cite{chen2019closer} normalizes the features and classification weights and calculates their cosine distances as the logits. \cite{mangla2020charting} pretrains with manifold mixup~\cite{verma2019manifold} to improve robustness. 
\cite{dhillon2020baseline} introduces a baseline for effectively and transductively finetuning. 
\cite{yang2021free} calibrates the training distribution of a few-shot task using statistics from the base set. \cite{tian2020rethinking} explores learning a general representation of base classes in supervised or self-supervised ways with self-distillation.

\textbf{Self-supervised learning.} We stress that in pFSL the pretrained model is unsupervised, for which the reason is to be explained in Sec.~\ref{sec:problem}. Self-supervised learning (SSL) is the mainstream approach to train a deep net in the unsupervised manner. Popular SSL methods, whether applied to traditional the ResNet~\cite{he2016deep} models (\eg, BYOL~\cite{grill2020bootstrap}, MoCo~\cite{MOCO} and SwAV~\cite{caron2020unsupervised}) or Vision Transformers (ViT)~\cite{dosovitskiy2020image} (\eg, MSN~\cite{assran2022masked}, DINO~\cite{caron2021emerging} and MoCov3~\cite{chen2021empirical}), all produce pretrained models based on large-scale datasets. In this paper, we utilize such models as our prior knowledge in FSL.

\section{pFSL: Practical Few-Shot Learning}\label{sec:problem}

In traditional FSL, the training dataset is split into 1) the \emph{base set} with $N_b$ labeled images, denoted as $D_b=\{(x^b_i, y^b_i)\}_{i=1}^{N_b}$ where $y^b_i \in \mathcal Y_b$ is the label of instance $x^b_i$ and $\mathcal Y_b$ is the label space of $D_b$; 2) the \emph{novel set} with $N_n$ labeled images, denoted as $D_n=\{(x^n_j, y^n_j)\}_{j=1}^{N_n}$ where $y^n_j \in \mathcal Y_n$ is the label of $x^n_j$ and $\mathcal Y_n$ is the label space of $D_n$. Note that although $\mathcal Y_b \cap \mathcal Y_n = \varnothing$, the concepts in $\mathcal Y_b$ and $\mathcal Y_n$ are semantically similar to each other. Then, the task is to learn a recognizer generalizing well on unknown samples from $\mathcal Y_n$. By letting $N_n \ll N_b$, the training samples in $D_n$ are few-shot (\ie, few training examples per class). In traditional FSL, $|\mathcal Y_b|$ is relatively large but $|\mathcal Y_n|$ is small. For example, in most cases $|\mathcal Y_n|=5$, which is called a 5-way FSL. The evaluation of traditional FSL is a rather complicated task.

\subsection{The pFSL Formulation}

To make FSL simpler, more practical, and closer to human's few-shot learning capability, we propose a new practical few-shot learning (pFSL) paradigm, which is characterized by the following properties:
\squishlist
 \item \textbf{Removing the base set.} The prior knowledge obtained through pretraining on the base set is limited, as the base set only contains limited number of concepts. In some applications, collecting data sharing similar concepts with the novel set may be as difficult as collecting more samples for the novel set itself.
 \item \textbf{Pretrained model based on big data.} A model pretrained on a large training set containing a wide variety of concepts is analogous to the common-sense and domain knowledge our brains encode, which will be useful for few shot learning in diverse domains.
 \item \textbf{Many-way FSL.} In an $N$-way $k$-shot FSL, we seek $N$ to be large and $k$ to be small, \eg, using all the CUB~\cite{wah2011caltech} categories ($N=200$ categories and $k$ is 1 or 5). This makes pFSL much closer to real-world applications than traditional FSL.
 \item \textbf{Unsupervised pretraining.} In pFSL, we require the pretraining to be unsupervised. Since a large scale dataset (\eg, ImageNet) may contain the concept in few-shot learning (e.g., those birds in CUB), avoiding using the labels during pretraining is necessary to make the evaluation of FSL algorithm fairer.
\squishend
These properties distinguish pFSL from traditional FSL and some recent variants of it~\cite{hu2022pushing,mangla2020charting,chen2019closer,dhillon2020baseline,kim2022better}.

\subsection{Simplicity and Better Evaluation}

The pFSL framework is clearly \emph{simpler} than traditional FSL, and it leads to not only benefits in the training phase, but more importantly \emph{better evaluation} of FSL algorithms.

Traditional $N$-way $k$-shot FSL typical use $N=5$ and $k=1,5$, which inevitably leads to unreliable estimation of the test accuracy. Hence, the common practice is to sample a large number of $n$ episodes ($n \ge 500$), find the accuracy in each episode, and report the average accuracy $\mu$ along with its 95\% confidence interval $Z_{95\%}$. Note that with $n$ episodes, the standard deviation $\sigma$ of the accuracy within the $n$ episodes is $\sigma=0.51\sqrt{n}Z_{95\%}$. For example, a typical result is: average accuracy 64.93\%, 95\% confidence interval 0.18\%, but the standard deviation of accuracy is 9.18\%~\cite{fu2022acsr} (in which $n=10000$). In short, one \emph{single} evaluation of traditional FSL \emph{requires training and testing for $n$ times ($n\ge 500$)} and \emph{the evaluation results are still unreliable}. Because of the complicated setup, both learning and evaluation of traditional FSL algorithms are not only costly, but often carried out in slightly different ways, which renders the fair comparison even more difficult.

As a stark contrast, pFSL is much simpler and as the experimental results will show in Table~\ref{table:tab1_low_shot}, the estimate of average accuracy is reliable. The standard deviation is small (mostly $<1.0\%$). This reliable estimation comes from the fact that pSFL is many-way (i.e., $N$ is large). Hence, instead of running $n$ episodes, \emph{few episodes} (\eg, 3) is enough to evaluate pFSL, which not only means savings in computation, but also more reliable evaluation of algorithms.
 
\section{IbM2: Instance-based Max-margin}\label{sec:IbM2}

In pFSL, we have a pretrained model $\mathcal{M}$ (unsupervised trained on a large set) and a novel set. There is no base set $D_b$ any more, hence we will simply denote the novel set as $D=\{x_i,y_i\}_{i=1}^M$ (that is, ignoring the superscript $^b$). The label set of $D$ is $\mathcal{Y}$. In an $N$-way $k$-shot pFSL task, we have $|\mathcal{Y}|=N$ and there will be $k$ training examples in each of the $N$ novel classes, hence $M=kN$. Note that we expect $|\mathcal{Y}|=N$ to be large (\ie, many-way FSL) and $k$ to be small (\eg, 1 or 5).

\subsection{Generating Virtual Samples}

Since $k$ (the number of training images in each class) is still small but the number of classes ($N$) becomes much larger (\eg, from 5 in traditional FSL to 200 in pFSL), we expect that the challenge in learning a pFSL model is greater than that in traditional FSL. This is verified by our experiments (\cf the results in Table~\ref{table:tab1_low_shot} vs. those in Table~\ref{table:tab3_fsl}).

Virtual examples, or examples that are sampled based on the original training examples, have been repeatedly proven useful when the number of training examples is small, \eg, in traditional few-shot learning~\cite{yang2021free} or long-tailed recognition (where the tail classes have few training images)~\cite{he2021distill}.

The common idea in generating virtual samples is to make the virtual samples follow the underlying distributions of the classes. However, in few-shot learning \emph{this requirement is very difficult to entertain} even with the help of techniques such as distribution calibration (DC) in~\cite{yang2021free}, because there is only 1 or 5 examples per class.

In contrast, we do not try to sample from any underlying distribution, but try to achieve max-margin of the decision boundary, as illustrated in Fig.~\ref{fig:intro_sketch}. Specifically, for every training \emph{instance} $x_i$, we i.i.d. generate $R$ virtual samples $x_{i,r}$ ($1\le r \le R$) which distribute around the shell of the hypersphere centered at $x_i$ and whose radius is controlled by a parameter $\epsilon$. Note that the same $\epsilon$ is shared by all training examples $x_i$ ($1 \le i \le kN$).

Technically, let $z_i$ be the feature vector produced by the pretrained model $\mathcal{M}$: $z_i=\mathcal{M}(x_i)$ from a forward calculation. Then, a noise vector $\delta_{i,r}$ is randomly sampled from the standard multi-dimensional normal distribution $N(0,I_d)$, where $d$ is the length of $z_i$ and $I_d$ is a $d \times d$ identity matrix. By adding a scaled version of the noise vector to the $i$-th training example, we obtain $z_{i,r}$ (the $r$-th virtual example for $x_i$):
\begin{equation}
	z_{i,r} = z_i + \epsilon \delta_{i,r} \,. \label{eq:noise1}
\end{equation}
Its label $y_{i,r}$ is $y_i$ (the label of $x_i$) and $\epsilon$ is a positive number. 

Interesting properties exist in our virtual examples:
\squishlist
 \item Hypersphere for different training examples \emph{can overlap}, so long as they belong to the same class (\cf Fig.~\ref{fig:intro_sketch}).
 \item We require the \emph{virtual} examples to be correctly classified, or, the original training examples are \emph{discarded}.
 \item The virtual examples lie \emph{around the shell, not the interior} of the hypersphere. According to the Gaussian Annulus Theorem~\cite{AvrimBlum2020FoundationsOD}, almost all the probability of a high-dimensional spherical Gaussian with unit variance is concentrated in a thin annulus of width $O(1)$ with radius $\sqrt{d}$ when $d$ is large. Since $d$ is indeed large (\eg, $d=2048$), the virtual examples reside around the shell. Hence, if we maximize the radius parameter $\epsilon$ but require that the virtual examples are correctly classified, as Fig.~\ref{fig:intro_sketch} shows, we in essence achieve margin maximization in an \emph{instance-based} manner.
\squishend
Hence, the proposed method is named Instance-based Max-margin, abbreviated as IbM2.

\subsection{Ellipsoidal Noise Generation}

One issue with the above isotropic noise sampling is that it ignores the structural property of the training examples. For instance, if the ranges for two feature dimensions are $[-1,1]$ and $[-100,100]$, respectively, isotropic sampling is clearly unsuitable. 

\begin{figure}
	\centering
	\includegraphics[width=0.9\linewidth]{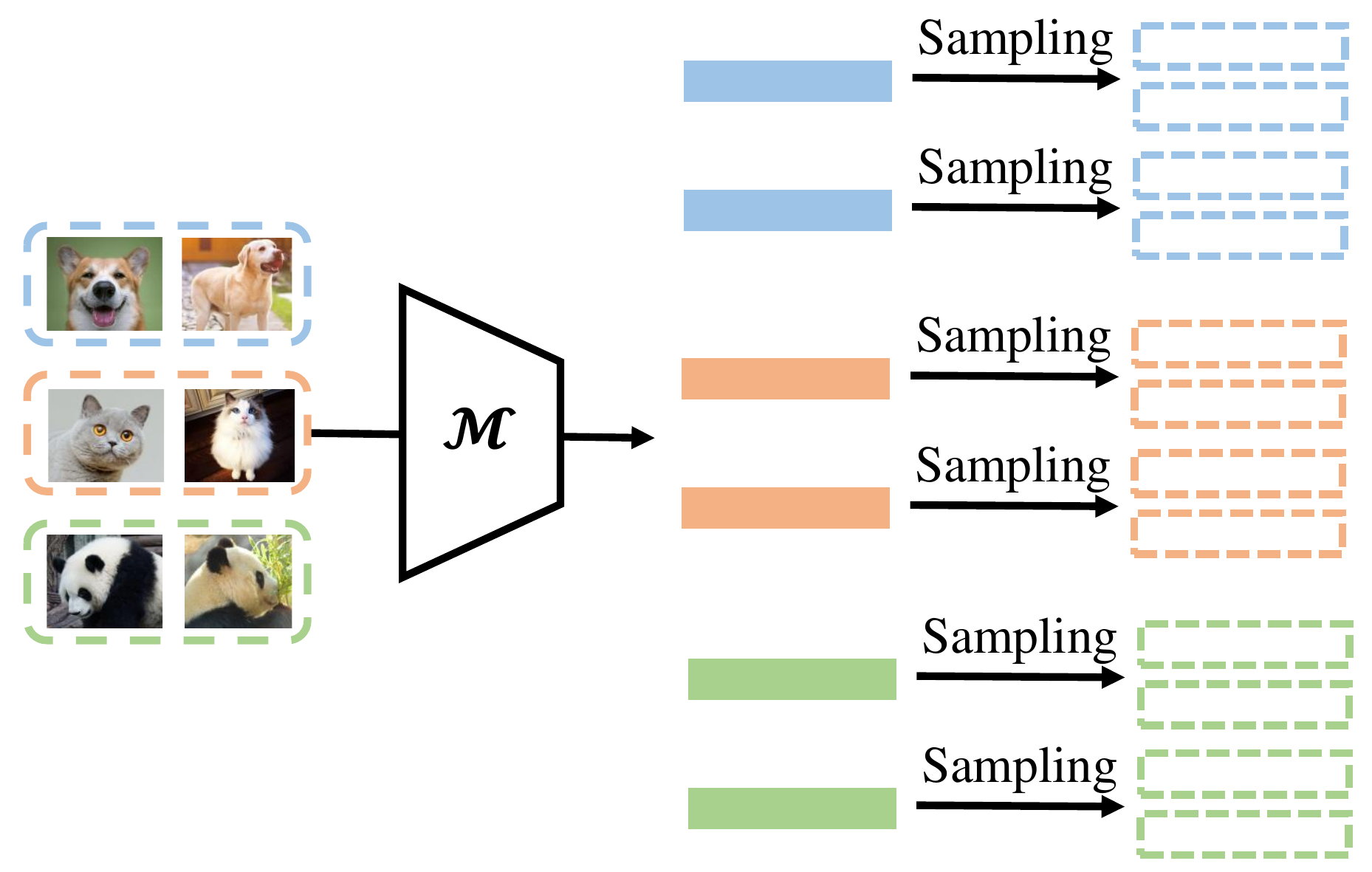}
	\caption{Virtual example generation. Original input images pass through the pretrained model $\mathcal{M}$ to get feature vectors (solid rectangles). Then, every feature vector i.i.d. samples $R$ virtual examples ($R=2$ here) using Eq.~\ref{eq:noise2}.}
	\label{fig:sampling}
\end{figure}

To overcome this drawback, we calculate a range vector $s=(s_1,s_2,\dots,s_d)$ using all the original training examples \emph{regardless of the class label}, where $s_i$ is the sample standard deviation for the $i$-th dimension. Then, the IbM2 virtual examples are generated as (replacing Eq.~\ref{eq:noise1})
\begin{equation}
	z_{i,r} = z_i + \epsilon (s\odot\delta_{i,r}) \,, \label{eq:noise2}
\end{equation}
where $\odot$ is element-wise multiplication. The final virtual example generation process is illustrated in Fig.~\ref{fig:sampling}. Note that estimating $s$ is dramatically easier than estimating the underlying distribution of every novel class.

It is worth noting that this ellipsoidal noise generation is \emph{unsupervised}, which distinguishes itself from Distribution calibration (DC)~\cite{yang2021free}. DC samples virtual examples based on an estimated covariance structure, which is estimated from the \emph{few} novel class training examples in a sample-wise manner and calibrated using the \emph{base set}. As aforementioned, we argue that it makes sense to get rid of the dependency on a base set, then DC is not applicable.

More importantly, the reliability of this estimation of a dense covariance matrix is understandably much lower than our estimation of the range indicator vector $s$---we estimate $d$ values using $k\times N$ examples, while DC estimates $d \times d$ values using only 1 example and the base set. The base set calibration will help DC, but clearly its estimation is still unreliable, as shown by experiments in~\cite{yang2021free,zhao2022exploring}

\subsection{Binary Search to Find the Largest Possible $\epsilon$}

Obviously we want $\epsilon$ to be as large as possible to achieve max-margin, under the condition that virtual examples are classified correctly. Hence, we design a simple binary search to find the optimal value for $\epsilon$.

For any given $\epsilon$ value, we can generate $kNR$ virtual examples $(x_{i,r},y_{i,r})$ ($1 \le i \le kN$, $1\le r \le R$) using Eq.~\ref{eq:noise2}. We denote the virtual training set for this particular value of $\epsilon$ as $D^\epsilon$. Then, we train a linear classifier with parameters $W$ to classify $D^\epsilon$ and obtain the training accuracy. If the training accuracy is too high (larger than a threshold $T$), $\epsilon$ is enlarged; otherwise $\epsilon$ is shrunk. The binary search maintains a search range, when the search range is tight enough (when the range is less than 0.05 in Algorithm~\ref{algo:FIMM}), the search terminates, and we have found the optimal value $\hat{\epsilon}$.

We have a few notes for finding $\hat{\epsilon}$:
\squishlist
 \item Unlike DC~\cite{yang2021free}, we do \emph{not} need a validation set.
 \item The threshold $T$ does not need to be 1 (100\%). In SVM learning, the slack variable trick allows some support vectors not to have max-margin. Similarly, we just need $T$ to be close to 1 (0.9 in our experiments).
\squishend
\begin{algorithm}[ht]
   \caption{Pseudo code for searching $\epsilon$}
   \label{algo:FIMM}
   \definecolor{codeblue}{rgb}{0.25,0.5,0.5}
   \lstset{
     backgroundcolor=\color{white},
     basicstyle=\fontsize{8pt}{8pt}\ttfamily\selectfont,
     breaklines=true,
     captionpos=b,
     commentstyle=\fontsize{8pt}{8pt}\color{codeblue},
     keywordstyle=\fontsize{8pt}{8pt}
   }
\begin{lstlisting}[language=python]
# Inputs:  
#     x  : training features of a few-shot task
#     y  : training labels   of a few-shot task
#     R  : sampling times for an instance
#     T  : accuracy threshold for searching
# Outputs: 
#     eps  : epsilon for sampling

left = 0.0
right = a large value
eps = right / 2

W = init_classifier
while True:
	acc = train_and_eval(W, x, y, eps, R)
	if acc > T:
		left = eps   # increase epsilon  
	else:
		right = eps  # decrease epsilon  
	eps = (left + right) / 2.0 
	if right - left < 0.05:
		break
\end{lstlisting}
\end{algorithm}

Finally, the IbM2 pipeline is simple: First, use Algorithm~\ref{algo:FIMM} to find $\hat{\epsilon}$. Second, generate the virtual dataset $D^{\hat{\epsilon}}$. And third, learn a linear classifier $W$. 

Note that the pretrained model $\mathcal{M}$ is frozen and will not be updated.

\section{Experimental Results}

We evaluate our IbM2 method on both setups: the proposed pFSL in many-way few-shot scenario, and also comparing IbM2 under the traditional FSL few-way setting with state-of-the-art FSL methods to further validate its effectiveness.

\subsection{Implementation Details}

For both the searching and training stages of IbM2, all features extracted by the backbone $\mathcal{M}$ were $L2$-normalized first. When learning the classification head (both during the binary search and after $\hat{\epsilon}$ was determined), we used Adam as the optimizer and the label smoothed cross-entropy loss~\cite{szegedy2016rethinking} as the learning objective. The pretrained model was trained using various self-supervised learning methods on the ImageNet-1K~\cite{russakovsky2015imagenet} training set.

During the binary search, for all the experiments we will report, we always set $R$ as 200, $T$ as 0.9 in pFSL and $0.999$ in the traditional FSL setting, except when we carried out ablation experiments on these hyperparameters.

Full implementation details are included in the appendix.

\subsection{Experiments in the pFSL Setting} \label{sec:experiment_on_pzfsl}

\textbf{Datasets and Evaluation Setup.} We explored two common many-way classification datasets, ImageNet-1K~\cite{russakovsky2015imagenet} and CUB-200-2011 (CUB)~\cite{wah2011caltech}. ImageNet-1K contains about 1.28 million training images from 1,000 classes and 50,000 images for evaluation. CUB is a fine-grained recognition dataset composed of 11,788 images belonging to 200 classes of birds, 5,994 for training and 5,794 for evaluation.

For both datasets, we randomly sampled the training sets for pFSL by selecting $k$ images from every class (\ie, $k$-shot) and using all the classes (1000 and 200, respectively) to form the novel set. $k$ is chosen from $\{1,2,3,4,5,8,16\}$. The evaluation was carried out on the full validation (for ImageNet) or test (for CUB) set.

Utilizing various self-supervised backbone network $\mathcal{M}$ pretrained on ImageNet, we compare two sets of results: one in which the linear classifier was obtained by using the original training set, the other by using IbM2's virtual examples. We report the top-1 accuracy and its standard deviation computed over 3 randomly sampled novel sets. 

\begin{table*}
	\centering
	\small
	\setlength{\tabcolsep}{2pt}
      \begin{tabular}{lcccccccccc}
         \hline
         \multirow{2}{*}{Dataset}                           & \multirow{2}{*}{\begin{tabular}[c]{@{}c@{}}Pretraining \\ Method\end{tabular}} & \multirow{2}{*}{Backbone} & \multirow{2}{*}{IbM2} & \multicolumn{7}{c}{Shot per Class}                                                                                                  \\
                                                            &                                                                                &                               &                       & 1                                       & 2                             & 3                            & 4                                 & 5                                & 8                                         & 16                                \\ \hline
         \hline                                             
         \multicolumn{1}{l|}{\multirow{14}{*}{ImageNet-1K}} & \multicolumn{1}{l|}{\multirow{2}{*}{DINO~\cite{caron2021emerging}}}                                     & \multirow{2}{*}{ViT-S/16}     &                       & 39.2 $\pm$ 0.3                          & 49.2 $\pm$ 0.2                & 54.1 $\pm$ 0.4              & 56.9 $\pm$ 0.2                     & 58.6 $\pm$ 0.1                   & 62.0 $\pm$ 0.2                            & 65.5 $\pm$ 0.2                    \\
         \multicolumn{1}{l|}{}                              & \multicolumn{1}{l|}{}                                                          &                               & $\checkmark$          & 39.3 $\pm$ 0.3                          & 49.5 $\pm$ 0.3                & \textbf{54.7 $\pm$ 0.4}     & \textbf{57.7 $\pm$ 0.1}            & \textbf{59.3 $\pm$ 0.2}          & 62.4 $\pm$ 0.2                            & 65.9 $\pm$ 0.1                    \\ \cline{2-11} 
         \multicolumn{1}{l|}{}                              & \multicolumn{1}{l|}{\multirow{2}{*}{MoCov3~\cite{chen2021empirical}}}                                   & \multirow{2}{*}{ViT-S/16}     &                       & 33.1 $\pm$ 0.6                          & 42.4 $\pm$ 0.3                & 47.2 $\pm$ 0.4              & 49.8 $\pm$ 0.4                     & 51.6 $\pm$ 0.3                   & 55.2 $\pm$ 0.2                            & 59.2 $\pm$ 0.3                    \\
         \multicolumn{1}{l|}{}                              & \multicolumn{1}{l|}{}                                                          &                               & $\checkmark$          & \textbf{34.4 $\pm$ 0.6}                 & \textbf{43.8 $\pm$ 0.3}       & \textbf{48.7 $\pm$ 0.4}     & \textbf{51.5 $\pm$ 0.3}            & \textbf{53.0 $\pm$ 0.2}          & \textbf{56.2 $\pm$ 0.1}                   & \textbf{60.3 $\pm$ 0.2}           \\ \cline{2-11} 
         \multicolumn{1}{l|}{}                              & \multicolumn{1}{l|}{\multirow{6}{*}{MSN\cite{assran2022masked}}}                                      & \multirow{2}{*}{ViT-S/16}     &                       & 47.9 $\pm$ 0.1                          & 56.2 $\pm$ 0.4                & 59.8 $\pm$ 0.3              & 61.7 $\pm$ 0.1                     & 62.8 $\pm$ 0.2                   & 65.3 $\pm$ 0.3                            & 68.1 $\pm$ 0.1                    \\
         \multicolumn{1}{l|}{}                              & \multicolumn{1}{l|}{}                                                          &                               & $\checkmark$          & 47.9 $\pm$ 0.1                          & 56.6 $\pm$ 0.5                & \textbf{60.5 $\pm$ 0.2}     & \textbf{62.5 $\pm$ 0.1}            & \textbf{63.6 $\pm$ 0.2}          & \textbf{66.0 $\pm$ 0.3}                   & 68.5 $\pm$ 0.1                    \\ \cline{3-11} 
         \multicolumn{1}{l|}{}                              & \multicolumn{1}{l|}{}                                                          & \multirow{2}{*}{ViT-B/4}      &                       & 53.9 $\pm$ 0.2                          & 64.5 $\pm$ 0.4                & 68.9 $\pm$ 0.3              & 70.9 $\pm$ 0.2                     & 72.3 $\pm$ 0.3                   & 74.1 $\pm$ 0.1                            & 75.7 $\pm$ 0.1                    \\
         \multicolumn{1}{l|}{}                              & \multicolumn{1}{l|}{}                                                          &                               & $\checkmark$          & 54.3 $\pm$ 0.1                          & \textbf{65.2 $\pm$ 0.6}       & \textbf{69.6 $\pm$ 0.2}     & \textbf{71.5 $\pm$ 0.2}            & 72.7 $\pm$ 0.3                   & \textbf{74.7 $\pm$ 0.0}                   & \textbf{76.4 $\pm$ 0.1}            \\ \cline{3-11} 
         \multicolumn{1}{l|}{}                              & \multicolumn{1}{l|}{}                                                          & \multirow{2}{*}{ViT-L/7}      &                       & 57.8 $\pm$ 0.3                          & 66.6 $\pm$ 0.5                & 69.9 $\pm$ 0.5              & 71.7 $\pm$ 0.4                     & 72.3 $\pm$ 0.2                   & 73.8 $\pm$ 0.1                            & 75.3 $\pm$ 0.1                    \\
         \multicolumn{1}{l|}{}                              & \multicolumn{1}{l|}{}                                                          &                               & $\checkmark$          & 58.2 $\pm$ 0.4                          & 66.9 $\pm$ 0.5                & 70.2 $\pm$ 0.6              & 71.9 $\pm$ 0.5                     & 72.7 $\pm$ 0.2                   & \textbf{74.3 $\pm$ 0.1}                   & \textbf{76.1 $\pm$ 0.0}            \\ \cline{2-11} 
         \multicolumn{1}{l|}{}                              & \multicolumn{1}{l|}{\multirow{2}{*}{SimCLR~\cite{chen2020simple}}}                                   & \multirow{2}{*}{ResNet50}     &                       & 24.2 $\pm$ 0.3                          & 33.3 $\pm$ 0.1                & 38.7 $\pm$ 0.4              & 41.6 $\pm$ 0.2                     & 43.5 $\pm$ 0.0                   & 47.2 $\pm$ 0.1                            & 51.9 $\pm$ 0.1                    \\
         \multicolumn{1}{l|}{}                              & \multicolumn{1}{l|}{}                                                          &                               & $\checkmark$          & 24.2 $\pm$ 0.3                          & 33.4 $\pm$ 0.2                & 39.0 $\pm$ 0.4              & 42.0 $\pm$ 0.3                     & \textbf{44.2 $\pm$ 0.1}          & \textbf{48.0 $\pm$ 0.0}                   & \textbf{52.7 $\pm$ 0.1}            \\ \cline{2-11} 
         \multicolumn{1}{l|}{}                              & \multicolumn{1}{l|}{\multirow{2}{*}{BYOL~\cite{grill2020bootstrap}}}                                     & \multirow{2}{*}{ResNet50}     &                       & 27.7 $\pm$ 0.2                          & 37.1 $\pm$ 0.1                & 42.5 $\pm$ 0.4              & 45.9 $\pm$ 0.3                     & 48.0 $\pm$ 0.1                   & 51.8 $\pm$ 0.1                            & 57.1 $\pm$ 0.1                    \\
         \multicolumn{1}{l|}{}                              & \multicolumn{1}{l|}{}                                                          &                               & $\checkmark$          & 27.6 $\pm$ 0.2                          & 37.5 $\pm$ 0.1                & \textbf{43.3 $\pm$ 0.4}     & \textbf{46.9 $\pm$ 0.2}            & \textbf{49.1 $\pm$ 0.1}          & \textbf{53.3 $\pm$ 0.2}                   & \textbf{58.0 $\pm$ 0.1}            \\ \hline
         \hline
         \multicolumn{1}{l|}{\multirow{8}{*}{CUB}}          & \multicolumn{1}{l|}{\multirow{2}{*}{DINO~\cite{caron2021emerging}}}                                     & \multirow{2}{*}{ViT-S/16}     &                       & 35.9 $\pm$ 1.4                          & 49.6 $\pm$ 0.3                & 57.4 $\pm$ 1.0              & 61.4 $\pm$ 0.8                     & 65.9 $\pm$ 0.9                   & 71.8 $\pm$ 0.6                            & 78.2 $\pm$ 0.1                    \\
         \multicolumn{1}{l|}{}                              & \multicolumn{1}{l|}{}                                                          &                               & $\checkmark$          & 36.2 $\pm$ 1.4                          & \textbf{50.2 $\pm$ 0.6}       & \textbf{57.9 $\pm$ 1.0}     & \textbf{62.4 $\pm$ 0.5}            & \textbf{66.7 $\pm$ 0.9}          & \textbf{72.6 $\pm$ 0.8}                   & \textbf{79.0 $\pm$ 0.1}            \\ \cline{2-11} 
         \multicolumn{1}{l|}{}                              & \multicolumn{1}{l|}{\multirow{2}{*}{MoCov3~\cite{chen2021empirical}}}                                   & \multirow{2}{*}{ViT-S/16}     &                       & 18.5 $\pm$ 0.7                          & 27.2 $\pm$ 0.2                & 35.5 $\pm$ 1.0              & 40.0 $\pm$ 0.6                     & 45.3 $\pm$ 0.9                   & 54.1 $\pm$ 0.4                            & 65.3 $\pm$ 0.3                    \\
         \multicolumn{1}{l|}{}                              & \multicolumn{1}{l|}{}                                                          &                               & $\checkmark$          & \textbf{19.5 $\pm$ 0.4}                 & 27.6 $\pm$ 0.2                & 35.6 $\pm$ 0.7              & 40.2 $\pm$ 0.4                     & \textbf{46.2 $\pm$ 1.1}          & \textbf{55.9 $\pm$ 0.4}                   & \textbf{66.8 $\pm$ 0.5}            \\ \cline{2-11} 
         \multicolumn{1}{l|}{}                              & \multicolumn{1}{l|}{\multirow{2}{*}{MSN~\cite{assran2022masked}}}                                      & \multirow{2}{*}{ViT-L/7}      &                       & 37.5 $\pm$ 1.1                          & 49.4 $\pm$ 0.4                & 58.8 $\pm$ 0.8              & 62.7 $\pm$ 0.9                     & 67.2 $\pm$ 0.3                   & 73.8 $\pm$ 0.8                            & 80.4 $\pm$ 0.2                    \\
         \multicolumn{1}{l|}{}                              & \multicolumn{1}{l|}{}                                                          &                               & $\checkmark$          & 37.7 $\pm$ 1.2                          & \textbf{50.1 $\pm$ 0.5}       & 59.0 $\pm$ 0.8              & \textbf{63.6 $\pm$ 0.5}            & \textbf{67.9 $\pm$ 0.7}          & \textbf{74.5 $\pm$ 0.5}                   & \textbf{81.2 $\pm$ 0.1}            \\ \cline{2-11} 
         \multicolumn{1}{l|}{}                              & \multicolumn{1}{l|}{\multirow{2}{*}{BYOL~\cite{grill2020bootstrap}}}                                     & \multirow{2}{*}{ResNet50}     &                       & 15.1 $\pm$ 0.8                          & 21.3 $\pm$ 0.6                & 28.4 $\pm$ 0.2              & 32.3 $\pm$ 0.2                     & 35.0 $\pm$ 1.1                   & 44.1 $\pm$ 0.1                            & 55.1 $\pm$ 0.5                    \\
         \multicolumn{1}{l|}{}                              & \multicolumn{1}{l|}{}                                                          &                               & $\checkmark$          & \textbf{16.4 $\pm$ 0.9}                 & \textbf{22.6 $\pm$ 0.6}       & \textbf{30.1 $\pm$ 0.6}     & \textbf{34.7 $\pm$ 0.5}            & \textbf{37.1 $\pm$ 1.3}          & \textbf{45.0 $\pm$ 0.4}                   & \textbf{57.6 $\pm$ 0.7}            \\ \hline
      \end{tabular}
   \caption{Average of top-1 accuracy (\%) with standard deviation across 3 random subsets on ImageNet-1K and CUB. A $\checkmark$ in the column IbM2 denotes the proposed IbM2 method, or the baseline method if it is blank. Those IbM2 results that are at least 0.5\% higher than the baseline are shown in boldface.}
   \label{table:tab1_low_shot}
\end{table*}

\textbf{Main Results.} As the results in Table~\ref{table:tab1_low_shot} suggest, we evaluated on two types of backbones: Vision Transformer (ViT)~\cite{dosovitskiy2020image} and ResNet50~\cite{he2016deep}. The pretraining methods included DINO~\cite{caron2021emerging}, MoCov3~\cite{chen2021empirical}, MSN~\cite{assran2022masked}, SimCLR~\cite{chen2020simple} and BYOL~\cite{grill2020bootstrap}. Table~\ref{table:tab1_low_shot} shows that \emph{in almost all cases}, IbM2 benefited the few-shot learning process and improved the top-1 accuracy. 

When the number of shots ($k$) is very small ($\le 2$), the improvement of IbM2 over the baseline is roughly $0.5\%$ on average. However, as the number of training shots goes larger, it is clear that the level of accuracy improvement increases gradually from $\sim 0.5\%$ to $\sim 1.0\%$, even $> 2\%$ in some cases.

A useful observation from these results is that in pFSL, both the baseline and IbM2 have small standard deviations, that is, more \emph{robust in evaluation}. Hence, we do not need 500 episodes in pFSL any longer---3 is enough.

\begin{table}
   \centering
   \small
   \renewcommand{\arraystretch}{1.04}
      \begin{tabular}{cccc}
         \hline
         \multirow{2}{*}{\begin{tabular}[c]{@{}c@{}}Pretraining \\ Method\end{tabular}} & \multirow{2}{*}{Backbone} & \multirow{2}{*}{IbM2} & \multirow{2}{*}{Top 1} \\
                                                                                        &                               &                       &                        \\ \hline
         \hline
         \multicolumn{1}{l|}{\multirow{4}{*}{DINO~\cite{caron2021emerging}}}                                     & \multirow{2}{*}{ViT-S/8}      &                       & 70.1                   \\
         \multicolumn{1}{l|}{}                                                          &                               & $\checkmark$          & 70.6                   \\ \cline{2-4} 
         \multicolumn{1}{l|}{}                                                          & \multirow{2}{*}{ViT-S/16}     &                       & 64.7                   \\
         \multicolumn{1}{l|}{}                                                          &                               & $\checkmark$          & 65.2                   \\ \hline
         \multicolumn{1}{l|}{\multirow{2}{*}{MoCov3~\cite{chen2021empirical}}}                                   & \multirow{2}{*}{ViT-S/16}     &                       & 58.1                   \\
         \multicolumn{1}{l|}{}                                                          &                               & $\checkmark$          & 59.1                   \\ \hline
         \multicolumn{1}{l|}{\multirow{6}{*}{MSN~\cite{assran2022masked}}}                                      & \multirow{2}{*}{ViT-S/16}     &                       & 67.3                   \\
         \multicolumn{1}{l|}{}                                                          &                               & $\checkmark$          & 68.0                   \\ \cline{2-4} 
         \multicolumn{1}{l|}{}                                                          & \multirow{2}{*}{ViT-B/4}      &                       & 75.4                   \\
         \multicolumn{1}{l|}{}                                                          &                               & $\checkmark$          & 76.1                   \\ \cline{2-4} 
         \multicolumn{1}{l|}{}                                                          & \multirow{2}{*}{ViT-L/7}      &                       & 74.8                   \\
         \multicolumn{1}{l|}{}                                                          &                               & $\checkmark$          & 75.6                   \\ \hline
         \multicolumn{1}{l|}{\multirow{2}{*}{SimCLR~\cite{chen2020simple}}}                                   & \multirow{2}{*}{ResNet50}     &                       & 50.5                   \\
         \multicolumn{1}{l|}{}                                                          &                               & $\checkmark$          & 51.3                   \\ \hline
         \multicolumn{1}{l|}{\multirow{2}{*}{BYOL~\cite{grill2020bootstrap}}}                                     & \multirow{2}{*}{ResNet50}     &                       & 55.7                   \\
         \multicolumn{1}{l|}{}                                                          &                               & $\checkmark$          & 56.9                   \\ \hline
         \end{tabular}
   \caption{Top-1 accuracy (\%) of 1\% ImageNet-1K semi-supervised learning. The training set contains on average 12 labeled training samples per category.}
   \label{table:tab2_1pt}
\end{table}

\textbf{Semi-supervised learning}. In the self-supervised learning literature, it is a common practice to use ImageNet-1K training images to learn a backbone in the unsupervised manner, then use a small portion (\eg, 1\%) of the training data now with labels to train a classifier. This semi-supervised learning task can also be viewed as in our pFSL setting. Hence, we also report the 1\% ImageNet-1K semi-supervised learning (on average 12 labeled training images per class) results in Table~\ref{table:tab2_1pt}. IbM2 consistently improved various self-supervised models and backbone architectures \emph{in all cases}, with 0.5\% to 1.2\% top-1 accuracy increase.

\begin{table*}
   \centering
   \small
   \setlength{\tabcolsep}{2pt}
      \begin{tabular}{l|l|lc|ccccc|ccccc}
         \hline
         \multirow{2}{*}{Dataset}       & \multirow{2}{*}{\begin{tabular}[c]{@{}l@{}}Pretraining \&\\ Meta-training\end{tabular}} & \multirow{2}{*}{Backbone}  & \multirow{2}{*}{IbM2} & \multicolumn{5}{c|}{1-shot}                                                         & \multicolumn{5}{c}{5-shot}                                                          \\
                                        &                                                                                       &                            &                       & $\text{ACC}_m$ & $\sigma$ & $\text{ACC}_1$ & $\text{ACC}_{10}$ & $\text{ACC}_{100}$ & $\text{ACC}_m$ & $\sigma$ & $\text{ACC}_1$ & $\text{ACC}_{10}$ & $\text{ACC}_{100}$ \\ \hline
         \hline
         \multirow{10}{*}{\textit{mini}-ImageNet} & \multirow{6}{*}{PMF~\cite{hu2022pushing}}                                   & \multirow{2}{*}{ViT-S/16}  &                       & 93.4          & 6.1     & 61.3          & 72.4             & 83.4              & 98.4          & 1.8     & 87.7          & 91.4             & 95.6              \\
                                        &                                                                                       &                            & $\checkmark$          & 94.4          & 5.6     & 63.2          & 74.7             & 85.1              & 98.6          & 1.8     & 86.1          & 91.3             & 95.9              \\ \cline{3-14} 
                                        &                                                                                       & \multirow{2}{*}{ViT-B/16}  &                       & 94.9          & 5.4     & 66.9          & 75.5             & 86.0              & 98.8          & 1.6     & 89.1          & 92.9             & 96.3              \\
                                        &                                                                                       &                            & $\checkmark$          & 95.6          & 5.1     & 69.6          & 77.6             & 87.1              & 98.9          & 1.6     & 89.3          & 93.0             & 96.3              \\ \cline{3-14} 
                                        &                                                                                       & \multirow{2}{*}{ResNet50}  & \multicolumn{1}{l|}{} & 94.9          & 5.3     & 67.2          & 75.2             & 86.3              & 98.5          & 1.7     & 89.3          & 91.9             & 95.8              \\
                                        &                                                                                       &                            & $\checkmark$          & 95.3          & 5.2     & 66.1          & 74.6             & 86.9              & 98.8          & 1.6     & 89.3          & 92.7             & 96.3              \\ \cline{2-14} 
                                        & \multirow{2}{*}{$S2M2_R$~\cite{mangla2020charting}}                                   & \multirow{2}{*}{WRN-28-10} & \multicolumn{1}{l|}{} & 65.4          & 10.0    & 32.0          & 39.3             & 50.7              & 82.3          & 7.1     & 52.0          & 62.5             & 71.8              \\
                                        &                                                                                       &                            & $\checkmark$          & 65.8          & 10.0    & 32.5          & 40.5             & 51.1              & 82.9          & 7.0     & 53.3          & 63.3             & 72.3              \\ \cline{2-14} 
                                        & \multirow{2}{*}{Meta-Baseline~\cite{chen2021meta}}                                    & \multirow{2}{*}{ResNet12}  & \multicolumn{1}{l|}{} & 62.9          & 10.4    & 26.7          & 34.3             & 47.4              & 79.0          & 7.5     & 54.7          & 58.0             & 67.9              \\
                                        &                                                                                       &                            & $\checkmark$          & 63.0          & 10.2    & 32.3          & 35.6             & 47.9              & 79.5          & 7.3     & 54.9          & 58.5             & 68.7              \\ \hline
         \hline
         \multirow{10}{*}{CIFAR-FS}     & \multirow{6}{*}{PMF~\cite{hu2022pushing}}                                             & \multirow{2}{*}{ViT-S/16}  &                       & 87.9          & 8.3     & 51.7          & 61.7             & 74.8              & 95.5          & 4.2     & 74.7          & 81.8             & 88.5              \\
                                        &                                                                                       &                            & $\checkmark$          & 89.0          & 8.1     & 59.2          & 64.1             & 75.9              & 95.7          & 4.2     & 73.3          & 81.7             & 88.7              \\ \cline{3-14} 
                                        &                                                                                       & \multirow{2}{*}{ViT-B/16}  &                       & 89.8          & 8.1     & 55.2          & 64.8             & 76.8              & 96.0          & 4.1     & 73.1          & 82.2             & 89.2              \\
                                        &                                                                                       &                            & $\checkmark$          & 90.3          & 8.0     & 58.9          & 65.5             & 77.4              & 96.0          & 4.1     & 74.1          & 81.8             & 89.2              \\ \cline{3-14} 
                                        &                                                                                       & \multirow{2}{*}{ResNet50}  & \multicolumn{1}{l|}{} & 81.7          & 10.4    & 35.2          & 51.6             & 65.8              & 91.3          & 5.5     & 71.2          & 75.1             & 82.7              \\
                                        &                                                                                       &                            & $\checkmark$          & 82.3          & 10.3    & 36.3          & 51.1             & 66.5              & 91.6          & 5.3     & 73.3          & 76.0             & 83.2              \\ \cline{2-14} 
                                        & \multirow{2}{*}{$S2M2_R$~\cite{mangla2020charting}}                                   & \multirow{2}{*}{WRN-28-10} & \multicolumn{1}{l|}{} & 75.2          & 10.8    & 44.0          & 46.7             & 59.2              & 87.7          & 7.1     & 54.7          & 67.9             & 76.8              \\
                                        &                                                                                       &                            & $\checkmark$          & 75.5          & 10.8    & 42.4          & 45.5             & 59.2              & 87.7          & 7.0     & 54.9          & 68.1             & 77.0              \\ \cline{2-14} 
                                        & \multirow{2}{*}{Meta-Baseline~\cite{chen2021meta}}                                    & \multirow{2}{*}{ResNet12}  & \multicolumn{1}{l|}{} & 72.7          & 11.6    & 32.0          & 42.9             & 56.1              & 84.8          & 7.5     & 54.7          & 65.1             & 73.6              \\
                                        &                                                                                       &                            & $\checkmark$          & 72.3          & 11.5    & 32.8          & 42.3             & 55.7              & 85.1          & 7.5     & 56.3          & 65.4             & 74.1              \\ \hline
         \end{tabular}
   \caption{Results of 5-way classification on \textit{mini}-ImageNet and CIFAR-FS in the traditional few-shot learning setting. For the baseline methods, we obtain the results from their respective official implementations.}
   \label{table:tab3_fsl}
\end{table*}

\subsection{Experiments in the Traditional FSL Setting}

\textbf{Datasets and Evaluation Setup.} We evaluated IbM2 in the traditional FSL setup, too. We conducted experiments on two standard benchmark datasets, \emph{mini}-ImageNet~\cite{vinyals2016matching} and CIFAR-FS~\cite{bertinetto2018meta}. \emph{mini}-ImageNet consists of 100 categories selected from ImageNet-1K, which are further split into 64 base, 16 val and 20 novel categories according to~\cite{ravi2016optimization}. CIFAR-FS is created by randomly shuffling the 100 categories of CIFAR-100~\cite{cifar_100} into 64 base, 16 val, 20 novel categories. Note that in order to perform traditional few-shot learning, only the novel split of these two datasets is required to sample many 5-way 1/5-shot episodes. We report the metrics mentioned in~\cite{fu2022acsr}: the average ($\text{ACC}_m$), worst-case ($\text{ACC}_1$), average of worst 10 ($\text{ACC}_{10}$), average of worst 100 ($\text{ACC}_{100}$) episodes' accuracy, and the standard deviation ($\sigma$) over 500 episodes for comprehensive evaluation.
To make the results more reliable, we average the value of each metric from 5 runs with different random seeds. We adopted PMF~\cite{hu2022pushing}, $S2M2_R$~\cite{mangla2020charting} and Meta-Baseline~\cite{chen2021meta} as the pretraining or meta-training approaches to obtain the backbone weights out of the base set.

\textbf{Main Results.} As shown in Table~\ref{table:tab3_fsl}, by plugging IbM2 in during classifier learning, the proposed IbM2 improved the average accuracy $\text{ACC}_m$ in all cases except only 1 result. And, the standard deviation $\sigma$ was reduced by IbM2 or remain unchanged in all cases. As~\cite{fu2022acsr} advocated, smaller $\sigma$ means the learning process is more stable.

\cite{fu2022acsr} also advocates the worst-case episodes accuracy is more important than the average accuracy among all episodes. Table~\ref{table:tab3_fsl} shows that IbM2 almost always leads to higher worst-case ($\text{ACC}_1$) or near-worst-case ($\text{ACC}_{10}$, $\text{ACC}_{100}$) accuracy. Furthermore, the average gain of $\text{ACC}_m$ is 0.4\%, which is far less than that of $\text{ACC}_1$ (1.3\%). As for $\text{ACC}_{10}$ and $\text{ACC}_{100}$, their gains are very similar, both around 0.5\%.
These observations demonstrate that IbM2 can boost the recognition accuracy of few-shot learning in almost every scenario, especially the worst case one.

At last, by comparing the numbers in Tables~\ref{table:tab1_low_shot} and~\ref{table:tab3_fsl}, pFSL is more \emph{challenging} than traditional FSL, which leads to more open questions to solve.

\subsection{Ablation Studies}

We performed ablation studies for IbM2 in the pFSL setting on ImageNet-1K.

\textbf{Instance-based vs. class-based max-margin.} IbM2 seeks max-margin through an instance-based manner. In this study, we compare this instance-based max-margin with the well-known support vector machine (SVM), which achieves max-margin in a class-based manner. We experimented with SVM using the linear kernel function in the LIBSVM~\cite{chang2011libsvm} software package. By iterating the value of SVM's regularization hyperparameter $C$ from the set $\{0.1, 1, 10, 100, 1000\}$, we found the best accuracy among them to compare with our IbM2 method. 

As shown in Table~\ref{table:tab4_svm}, the proposed IbM2 method outperforms SVM with the best $C$ by a significant margin in all cases. Specifically, the recognition accuracy improves $\sim$2\% on average, and even up to $\sim$4\% in the extremely scarce 1-shot case. These results demonstrate that our instance-based margin generated by IbM2 helps more to learn a robust classification boundary, especially when the training distribution is drastically shifted.

\begin{table}
   \centering
   \small
      \begin{tabular}{c|ccccccc}
         \hline
         \multirow{2}{*}{Method} & \multicolumn{7}{c}{Shot per Class}             \\
                                  & 1    & 2    & 3    & 4    & 5    & 8    & 16   \\ \hline
         SVM                      & 54.8 & 64.8 & 68.2 & 70.0 & 70.8 & 72.8 & 74.7 \\
         IbM2                     & 58.2 & 66.9 & 70.2 & 71.9 & 72.7 & 74.3 & 76.1 \\ \hline
         \end{tabular}
   \caption{Top-1 accuracy (\%) with different margin-based methods on ImageNet-1K classification with ViT-L/7 from MSN~\cite{assran2022masked} as the pretrained backbone.}
   \label{table:tab4_svm}
\end{table}

\begin{table}
   \centering
   \small
   
      \begin{tabular}{c|ccccccc}
         \hline
         \multirow{2}{*}{$R$} & \multicolumn{7}{c}{Shot per Class}                                                                                                                 \\
                            & 1                        & \multicolumn{1}{c}{2}    & 3    & 4    & \multicolumn{1}{c}{5}    & \multicolumn{1}{c}{8}    & \multicolumn{1}{c}{16}   \\ \hline
         1                  & 32.9                     & \multicolumn{1}{c}{42.7} & 47.8 & 50.8 & \multicolumn{1}{c}{52.5} & \multicolumn{1}{c}{55.9} & \multicolumn{1}{c}{60.1} \\
         10                 & 33.7                     & \multicolumn{1}{c}{43.6} & 48.5 & 51.4 & \multicolumn{1}{c}{52.9} & \multicolumn{1}{c}{56.2} & \multicolumn{1}{c}{60.3} \\
         50                 & \multicolumn{1}{l}{33.8} & 43.7                     & 48.6 & 51.5 & 53.0                     & 56.2                     & 60.3                     \\
         200                & \multicolumn{1}{l}{34.4} & 43.8                     & 48.7 & 51.5 & 53.0                     & 56.2                     & 60.3                     \\
         400                & \multicolumn{1}{l}{34.5} & 43.8                     & 48.6 & 51.5 & 53.0                     & 56.2                     & 60.3                     \\ \hline
         \end{tabular}
   \caption{Top-1 accuracy (\%) of classification with different $R$ on ImageNet-1K with ViT-S/16 from MoCov3~\cite{chen2021empirical} as the pretrained backbone.}
   \label{table:tab5_M}
\end{table}

\textbf{Sensitivity of $R$.} As described in Sec.~\ref{sec:IbM2}, one original training example is turned into $R$ virtual examples to form the training set $D^{\hat{\epsilon}}$. In this part, we study the effect of the hyperparameter $R$. As shown in Table~\ref{table:tab5_M}, when the training shots are highly limited, increasing $R$ significantly improves recognition accuracy.
As the number of training shots gets larger, the accuracy difference between a large $R$ ($R=400$) and a small one ($R=1$) gradually decreases from 1.6\% to 0.2\%.

When training samples are extremely scarce, it is difficult to model the correct Gaussian distribution with a few samples in a high-dimensional space. Therefore, increasing $R$ is necessary for a good estimation. When more training shots are available, the need for a large value of sampling times is reduced, hence the accuracy difference between different $R$ becomes smaller, too. Based on these results, we let $R=200$ in all our experiments.

\begin{table}
   \centering
   \small
   \setlength{\tabcolsep}{5pt}
      \begin{tabular}{c|ccccccc}
         \hline
         \multirow{2}{*}{\begin{tabular}[c]{@{}c@{}}Sampling \\ Schema\end{tabular}} & \multicolumn{7}{c}{Shot per Class}                                                                                                                 \\
                                                                                     & 1                        & 2                        & 3    & 4    & 5                        & 8                        & 16                       \\ \hline
         -                                                                           & 33.1                     & 42.4                     & 47.2 & 49.8 & 51.6                     & 55.2                     & 59.2                     \\
         Spherical                                                                    & 33.0                     & 42.1                     & 47.1 & 50.2 & 51.8                     & 55.3                     & 59.8                     \\
         Ellipsoidal                                                                 & \multicolumn{1}{l}{34.4} & \multicolumn{1}{l}{43.8} & 48.7 & 51.5 & \multicolumn{1}{l}{53.0} & \multicolumn{1}{l}{56.2} & \multicolumn{1}{l}{60.3} \\ \hline
         \end{tabular}
   \caption{Top-1 accuracy (\%) of classification with different sampling schemas on ImageNet-1K with ViT-S/16 from MoCov3~\cite{chen2021empirical} as the pretrained backbone. '-' denotes the simple baseline using only the original training examples.}
   \label{table:tab6_ellipsoidal}
\end{table}

\textbf{Ellipsoidal vs. isotropic sampling.} We have described two noise sampling strategies to generate virtual examples. The ellipsoidal one is preferred over the isotropic one, and is used in IbM2. In this part, we evaluate the effectiveness of this sampling schema by experimentally comparing these two sampling strategies. The isotropic sampling strategy is denoted as \emph{Spherical} in Table~\ref{table:tab6_ellipsoidal}.

As shown in Table~\ref{table:tab6_ellipsoidal}, the accuracy in the ellipsoidal sampling schema is consistently better than its spherical counterpart or the baseline. Moreover, in low-shot cases ($\le 3$), sampling with spherical Gaussian slightly degraded the recognition accuracy. The reason may be that the standard deviation of different channels calculated from training features varied a lot, thus simply regarding all channels as independent and identically distributed might make the sampled points significantly shifted from the underlying distribution. However, as the number of training samples increases, both isotropic and ellipsoidal sampling outperformed the baseline.

Based on these observations, we chose to adopt the ellipsoidal noise sampling (Eq.~\ref{eq:noise2}) in IbM2.

\textbf{On what classes can IbM2 help?} Finally, we study what classes will benefit from IbM2---will most or only a small portion of classes be improved by IbM2? After the training of the baseline and IbM2 finished, we calculated the  class-wise validation accuracy for every class on ImageNet-1K, as
\begin{equation}
   \text{ACC}^k = \frac{N_{cor}^k}{N_{cls}^k} \,,
   \label{eq:acc_class_wise}
\end{equation}
where $\text{ACC}^k$ is the per-class accuracy (recall) of the $k$-th class, $N_{cls}^k$ is the number of test samples from class $k$ ($N_{cls}^k=50$ in ImageNet-1K) and ${N_{cor}^k}$ is the number of correctly classified samples of class $k$. For every class, we obtain the difference of per-class accuracy (that of IbM2 minus that of the baseline). A positive difference is an accuracy gain and a negative gain is in fact an accuracy loss. 

\begin{figure}
   \centering
   \includegraphics[width=0.99\linewidth]{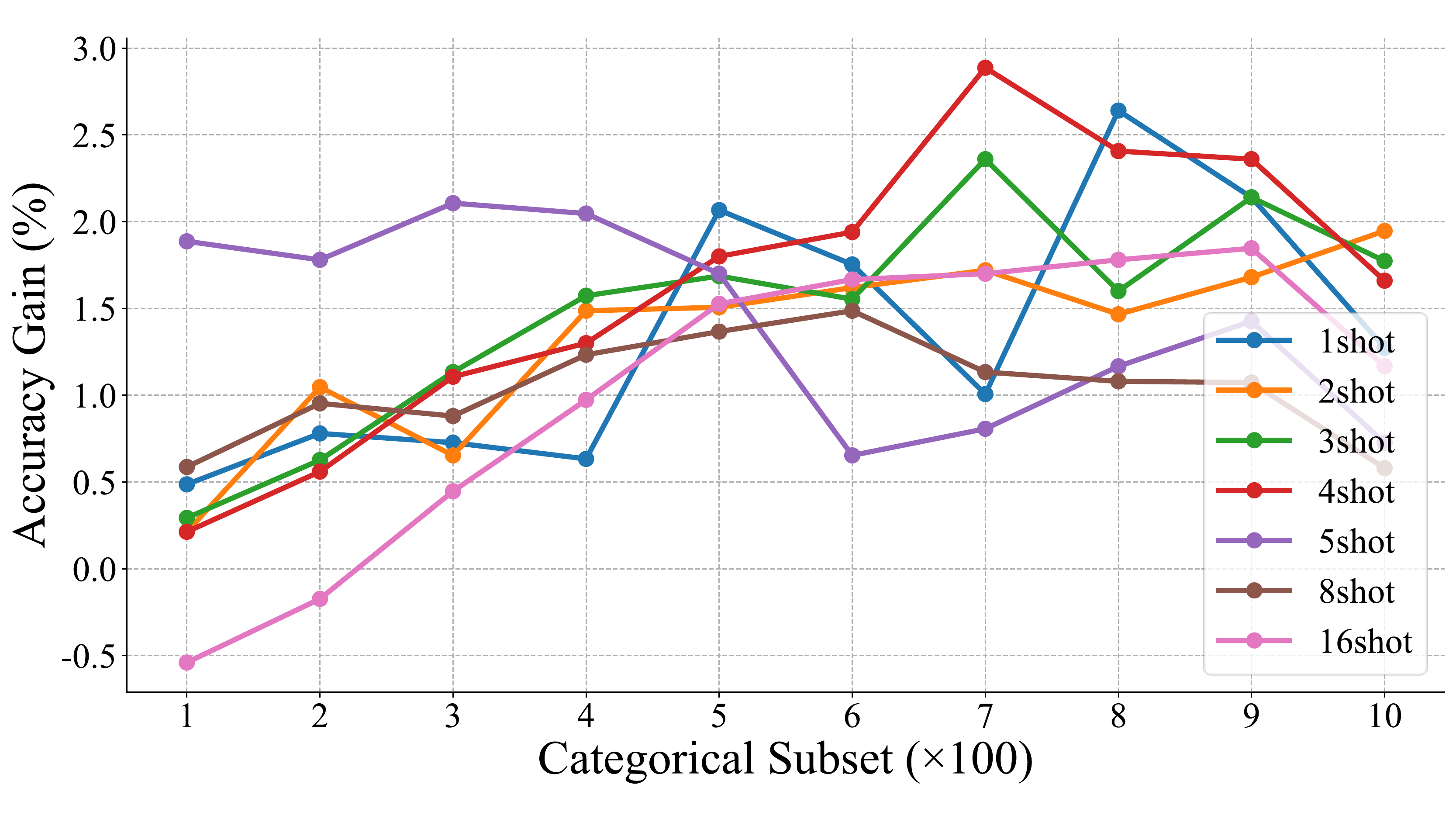}
   \caption{Categorical accuracy gain (\%) of IbM2 compared to baseline with ViT-S/16 pretrained by MoCov3~\cite{chen2021empirical} as the backbone.}
   \label{fig:acc_gain}
\end{figure}

We then sort the baseline's per-class accuracy in the ascending order. Following this order, we rearrange the per-class accuracy gains. The 1000 accuracy gains are divided into 10 histogram bins in the rearranged order, and inside each bin the 100 per-class accuracy gains are averaged to obtain the accuracy gain of that bin. Fig.~\ref{fig:acc_gain} plots the average accuracy gain in each bin.

We find that IbM2 improves average per-class accuracy in almost every histogram bin (\ie, is above the ``0.0'' horizon in the $y$-axis). More detailed numbers reveal that the recognition accuracy of more than 65\% categories out of 1000 is improved in \emph{ all} shots. Furthermore, Fig.~\ref{fig:acc_gain} shows a trend: the more accurate a class is, the higher gains IbM2 can achieve over the baseline. That is, when the task is too difficult for the baseline, IbM2 can help but its usefulness is restricted. But, when the baseline is already accurate enough (categorical subset 9 and later), the room for IbM2's further improvement is small again. It is the middle range that IbM2 is the most useful.

\section{Conclusions and Limitations}

In this paper, we advocated pFSL, a new practical few-shot learning setting. We also proposed IbM2, an instance-based max-margin method to improve few-shot learning. With the technological advancements, it is the time to upgrade the traditional FSL settings. We need a simpler, easier to evaluate, more challenging and more practical FSL setting, and the proposed pFSL task satisfies these requirements.

To face the challenges related to the scarcity of training examples, we proposed IbM2. Instead of maximizing the class-level margin, IbM2 is instance-based margin maximization. It achieved significant improvements in both pFSL and the traditional FSL settings consistently. IbM2 is simple and reliable. It introduces only 2 hyperparameters, and the default values of them worked well in experiments across different architectures and tasks.

As the experiments indicated, IbM2 works the best for the middle range in terms of task difficulty. In a $k$-shot task when $k$ is extremely small, there is still a lot of room for improvements. Furthermore, IbM2 freezes the backbone and only learns the linear classifier. In the future, we plan to further improve the very small shot cases (\eg, $k=1$) and better tune the backbone in few-shot learning.

{\small
\bibliographystyle{ieee_fullname}
\bibliography{egbib}

\begin{thebibliography}{10}\itemsep=-1pt

\bibitem{afrasiyabi2022matching}
Arman Afrasiyabi, Hugo Larochelle, Jean-François Lalonde, and Christian
  Gagné.
\newblock Matching feature sets for few-shot image classification.
\newblock In {\em IEEE/CVF Conference on Computer Vision and Pattern
  Recognition}, pages 9004--9014, 2022.

\bibitem{assran2022masked}
Mahmoud Assran, Mathilde Caron, Ishan Misra, Piotr Bojanowski, Florian Bordes,
  Pascal Vincent, Armand Joulin, Mike Rabbat, and Nicolas Ballas.
\newblock {Masked Siamese Networks} for label-efficient learning.
\newblock In {\em European Conference on Computer Vision}, volume 13691 of {\em
  LNCS}, page 456–473. Springer, 2022.

\bibitem{bertinetto2018meta}
Luca Bertinetto, Joao~F Henriques, Philip~HS Torr, and Andrea Vedaldi.
\newblock Meta-learning with differentiable closed-form solvers.
\newblock In {\em International Conference on Learning Representations}, pages
  1--15, 2019.

\bibitem{AvrimBlum2020FoundationsOD}
Avrim Blum, John Hopcroft, and Ravindran Kannan.
\newblock {\em Foundations of Data Science}.
\newblock Cambridge University Press, 2020.

\bibitem{caron2020unsupervised}
Mathilde Caron, Ishan Misra, Julien Mairal, Priya Goyal, Piotr Bojanowski, and
  Armand Joulin.
\newblock Unsupervised learning of visual features by contrasting cluster
  assignments.
\newblock In {\em Advances in Neural Information Processing Systems}, pages
  9912--9924, 2020.

\bibitem{caron2021emerging}
Mathilde Caron, Hugo Touvron, Ishan Misra, Hervé Jegou, Julien Mairal, Piotr
  Bojanowski, and Armand Joulin.
\newblock Emerging properties in self-supervised {Vision Transformers}.
\newblock In {\em IEEE/CVF International Conference on Computer Vision}, pages
  9630--9640, 2021.

\bibitem{chang2011libsvm}
Chih-Chung Chang and Chih-Jen Lin.
\newblock {LIBSVM}: A library for support vector machines.
\newblock {\em ACM Transactions on Intelligent Systems and Technology},
  2(3):1--27, 2011.

\bibitem{chen2020simple}
Ting Chen, Simon Kornblith, Mohammad Norouzi, and Geoffrey Hinton.
\newblock A simple framework for contrastive learning of visual
  representations.
\newblock In {\em International Conference on Machine Learning}, pages
  1597--1607, 2020.

\bibitem{chen2019closer}
Wei-Yu Chen, Yen-Cheng Liu, Zsolt Kira, Yu-Chiang~Frank Wang, and Jia-Bin
  Huang.
\newblock A closer look at few-shot classification.
\newblock In {\em International Conference on Learning Representations}, pages
  1--16, 2019.

\bibitem{chen2021empirical}
Xinlei Chen, Saining Xie, and Kaiming He.
\newblock An empirical study of training self-supervised {Vision Transformers}.
\newblock In {\em IEEE/CVF International Conference on Computer Vision}, pages
  9620--9629, 2021.

\bibitem{chen2021meta}
Yinbo Chen, Zhuang Liu, Huijuan Xu, Trevor Darrell, and Xiaolong Wang.
\newblock {Meta-Baseline}: Exploring simple meta-learning for few-shot
  learning.
\newblock In {\em IEEE/CVF International Conference on Computer Vision}, pages
  9042--9051, 2021.

\bibitem{dhillon2020baseline}
Guneet~S Dhillon, Pratik Chaudhari, Avinash Ravichandran, and Stefano Soatto.
\newblock A baseline for few-shot image classification.
\newblock In {\em International Conference on Learning Representations}, pages
  1--20, 2020.

\bibitem{dosovitskiy2020image}
Alexey Dosovitskiy, Lucas Beyer, Alexander Kolesnikov, Dirk Weissenborn,
  Xiaohua Zhai, Thomas Unterthiner, Mostafa Dehghani, Matthias Minderer, Georg
  Heigold, Sylvain Gelly, Jakob Uszkoreit, and Neil Houlsby.
\newblock An image is worth 16x16 words: {Transformers} for image recognition
  at scale.
\newblock In {\em International Conference on Learning Representations}, pages
  1--21, 2021.

\bibitem{finn2017maml}
Chelsea Finn, Pieter Abbeel, and Sergey Levine.
\newblock Model-agnostic meta-learning for fast adaptation of deep networks.
\newblock In {\em International Conference on Machine Learning}, page
  1126–1135, 2017.

\bibitem{finn2018probabilistic}
Chelsea Finn, Kelvin Xu, and Sergey Levine.
\newblock Probabilistic model-agnostic meta-learning.
\newblock In {\em Advances in Neural Information Processing Systems}, page
  9537–9548, 2018.

\bibitem{fu2022acsr}
Minghao Fu, Yun-Hao Cao, and Jianxin Wu.
\newblock Worst case matters for few-shot recognition.
\newblock In {\em European Conference on Computer Vision}, volume 13680 of {\em
  LNCS}, page 99–115. Springer, 2022.

\bibitem{grill2020bootstrap}
Jean-Bastien Grill, Florian Strub, Florent Altch\'{e}, Corentin Tallec,
  Pierre~H. Richemond, Elena Buchatskaya, Carl Doersch, Bernardo~Avila Pires,
  Zhaohan~Daniel Guo, Mohammad~Gheshlaghi Azar, Bilal Piot, Koray Kavukcuoglu,
  R\'{e}mi Munos, and Michal Valko.
\newblock {Bootstrap Your Own Latent} - a new approach to self-supervised
  learning.
\newblock In {\em Advances in Neural Information Processing Systems}, pages
  21271--21284, 2020.

\bibitem{MOCO}
Kaiming He, Haoqi Fan, Yuxin Wu, Saining Xie, and Ross Girshick.
\newblock {Momentum Contrast} for unsupervised visual representation learning.
\newblock In {\em IEEE/CVF Conference on Computer Vision and Pattern
  Recognition}, pages 9726--9735, 2020.

\bibitem{he2016deep}
Kaiming He, Xiangyu Zhang, Shaoqing Ren, and Jian Sun.
\newblock Deep residual learning for image recognition.
\newblock In {\em IEEE Conference on Computer Vision and Pattern Recognition},
  pages 770--778, 2016.

\bibitem{he2021distill}
Yin-Yin He, Jianxin Wu, and Xiu-Shen Wei.
\newblock Distilling virtual examples for long-tailed recognition.
\newblock In {\em IEEE/CVF International Conference on Computer Vision}, pages
  235--244, 2021.

\bibitem{hospedales2021meta_survey}
Timothy Hospedales, Antreas Antoniou, Paul Micaelli, and Amos Storkey.
\newblock Meta-learning in neural networks: A survey.
\newblock {\em IEEE Transactions on Pattern Analysis and Machine Intelligence},
  44(9):5149--5169, 2022.

\bibitem{hu2022pushing}
Shell~Xu Hu, Da Li, Jan Stühmer, Minyoung Kim, and Timothy~M. Hospedales.
\newblock Pushing the limits of simple pipelines for few-shot learning:
  External data and fine-tuning make a difference.
\newblock In {\em IEEE/CVF Conference on Computer Vision and Pattern
  Recognition}, pages 9058--9067, 2022.

\bibitem{kim2022better}
Seong-Woong Kim and Dong-Wan Choi.
\newblock Better generalized few-shot learning even without base data.
\newblock In {\em AAAI Conference on Artificial Intelligence}, 2023.

\bibitem{cifar_100}
Alex Krizhevsky and Geoffrey Hinton.
\newblock Learning multiple layers of features from tiny images.
\newblock Technical Report~0, University of Toronto, 2009.

\bibitem{lee2019meta}
Kwonjoon Lee, Subhransu Maji, Avinash Ravichandran, and Stefano Soatto.
\newblock Meta-learning with differentiable convex optimization.
\newblock In {\em IEEE/CVF Conference on Computer Vision and Pattern
  Recognition}, pages 10649--10657, 2019.

\bibitem{lu2022self}
Yuning Lu, Liangjian Wen, Jianzhuang Liu, Yajing Liu, and Xinmei Tian.
\newblock Self-supervision can be a good few-shot learner.
\newblock In {\em European Conference on Computer Vision}, volume 13679 of {\em
  LNCS}, pages 740--758. Springer, 2022.

\bibitem{luo2021rectifying}
Xu Luo, Longhui Wei, Liangjian Wen, Jinrong Yang, Lingxi Xie, Zenglin Xu, and
  Qi Tian.
\newblock Rectifying the shortcut learning of background for few-shot learning.
\newblock In {\em Advances in Neural Information Processing Systems}, pages
  13073--13085, 2021.

\bibitem{mangla2020charting}
Puneet Mangla, Mayank Singh, Abhishek Sinha, Nupur Kumari, Vineeth~N
  Balasubramanian, and Balaji Krishnamurthy.
\newblock Charting the right manifold: {Manifold Mixup} for few-shot learning.
\newblock In {\em IEEE Winter Conference on Applications of Computer Vision},
  pages 2207--2216, 2020.

\bibitem{ravi2016optimization}
Sachin Ravi and Hugo Larochelle.
\newblock Optimization as a model for few-shot learning.
\newblock In {\em International Conference on Learning Representations}, pages
  1--11, 2017.

\bibitem{russakovsky2015imagenet}
Olga Russakovsky, Jia Deng, Hao Su, Jonathan Krause, Sanjeev Satheesh, Sean Ma,
  Zhiheng Huang, Andrej Karpathy, Aditya Khosla, Michael Bernstein,
  Alexander~C. Berg, and Li Fei-Fei.
\newblock {ImageNet} large scale visual recognition challenge.
\newblock {\em International Journal of Computer Vision}, 115(3):211--252,
  2015.

\bibitem{rusu2019meta}
Andrei~A Rusu, Dushyant Rao, Jakub Sygnowski, Oriol Vinyals, Razvan Pascanu,
  Simon Osindero, and Raia Hadsell.
\newblock Meta-learning with latent embedding optimization.
\newblock In {\em International Conference on Learning Representations}, pages
  1--17, 2019.

\bibitem{snell2017prototypical}
Jake Snell, Kevin Swersky, and Richard Zemel.
\newblock {Prototypical Networks} for few-shot learning.
\newblock In {\em Advances in Neural Information Processing Systems}, page
  4080–4090, 2017.

\bibitem{su2020does}
Jong-Chyi Su, Subhransu Maji, and Bharath Hariharan.
\newblock When does self-supervision improve few-shot learning?
\newblock In {\em European Conference on Computer Vision}, volume 12352 of {\em
  LNCS}, pages 645--666. Springer, 2020.

\bibitem{sung2018learning}
Flood Sung, Yongxin Yang, Li Zhang, Tao Xiang, Philip~H.S. Torr, and Timothy~M.
  Hospedales.
\newblock Learning to compare: {Relation Network} for few-shot learning.
\newblock In {\em IEEE Conference on Computer Vision and Pattern Recognition},
  pages 1199--1208, 2018.

\bibitem{szegedy2016rethinking}
Christian Szegedy, Vincent Vanhoucke, Sergey Ioffe, Jon Shlens, and Zbigniew
  Wojna.
\newblock Rethinking the {Inception} architecture for computer vision.
\newblock In {\em IEEE Conference on Computer Vision and Pattern Recognition},
  pages 2818--2826, 2016.

\bibitem{tian2020rethinking}
Yonglong Tian, Yue Wang, Dilip Krishnan, Joshua~B Tenenbaum, and Phillip Isola.
\newblock Rethinking few-shot image classification: a good embedding is all you
  need?
\newblock In {\em European Conference on Computer Vision}, volume 12359 of {\em
  LNCS}, page 266–282. Springer, 2020.

\bibitem{verma2019manifold}
Vikas Verma, Alex Lamb, Christopher Beckham, Amir Najafi, Ioannis Mitliagkas,
  David Lopez-Paz, and Yoshua Bengio.
\newblock {Manifold Mixup}: Better representations by interpolating hidden
  states.
\newblock In {\em International Conference on Machine Learning}, pages
  6438--6447, 2019.

\bibitem{vinyals2016matching}
Oriol Vinyals, Charles Blundell, Timothy Lillicrap, Koray Kavukcuoglu, and Daan
  Wierstra.
\newblock {Matching Networks} for one shot learning.
\newblock In {\em Advances in Neural Information Processing Systems}, page
  3637–3645, 2016.

\bibitem{wah2011caltech}
Catherine Wah, Steve Branson, Peter Welinder, Pietro Perona, and Serge
  Belongie.
\newblock The {Caltech-UCSD} {Birds-200-2011} dataset.
\newblock Technical Report CNS-TR-2011-001, California Institute of Technology,
  2011.

\bibitem{wang2020generalizing}
Yaqing Wang, Quanming Yao, James~T. Kwok, and Lionel~M. Ni.
\newblock Generalizing from a few examples: A survey on few-shot learning.
\newblock {\em ACM Computing Surveys}, 53(3):1--34, 2020.

\bibitem{yang2021free}
Shuo Yang, Lu Liu, and Min Xu.
\newblock Free lunch for few-shot learning: Distribution calibration.
\newblock In {\em International Conference on Learning Representations}, pages
  1--13, 2021.

\bibitem{yang2022few}
Zhanyuan Yang, Jinghua Wang, and Yingying Zhu.
\newblock Few-shot classification with contrastive learning.
\newblock In {\em European Conference on Computer Vision}, volume 13680 of {\em
  LNCS}, page 293–309. Springer, 2022.

\bibitem{ye2020few}
Han-Jia Ye, Hexiang Hu, De-Chuan Zhan, and Fei Sha.
\newblock Few-shot learning via embedding adaptation with set-to-set functions.
\newblock In {\em IEEE/CVF Conference on Computer Vision and Pattern
  Recognition}, pages 8805--8814, 2020.

\bibitem{wrn_28_10}
Sergey Zagoruyko and Nikos Komodakis.
\newblock Wide residual networks.
\newblock In {\em British Machine Vision Conference}, pages 1--12, 2016.

\bibitem{zhang2022deepemd}
Chi Zhang, Yujun Cai, Guosheng Lin, and Chunhua Shen.
\newblock {DeepEMD}: Few-shot image classification with differentiable {Earth
  Mover}'s distance and structured classifiers.
\newblock In {\em IEEE/CVF Conference on Computer Vision and Pattern
  Recognition}, pages 12200--12210, 2020.

\bibitem{zhao2022exploring}
Zhiyuan Zhao, Qingjie Liu, and Yunhong Wang.
\newblock Exploring effective knowledge transfer for few-shot object detection.
\newblock In {\em ACM International Conference on Multimedia}, page
  6831–6839, 2022.

\end{thebibliography}
}

\clearpage
\appendix
\section*{Appendix}

In this appendix, we provide the full implementation details of experiments in the proposed pFSL or traditional FSL settings.

First, we describe the common details in all experiments. The features $z$ extracted by the backbone $\mathcal M$ were $L2$-normalized. Whenever a linear classifier was trained, we always used Adam as the optimizer whose learning rate was gradually decreased in a cosine scheduling, and the label smoothed cross-entropy loss~\cite{szegedy2016rethinking} was used as the learning objective.

\section{Setup for pFSL}

The pretrained model $\mathcal M$ was trained using various self-supervised learning methods~\cite{caron2021emerging,chen2021empirical,assran2022masked,chen2020simple,grill2020bootstrap} on the ImageNet-1K~\cite{russakovsky2015imagenet} training set. We obtained the pretrained models from their respective official implementations. To store the features extracted by the backbone $\mathcal M$ in advance for conveniently conducting experiments, images were resized to 256 pixels along the shorter side using bicubic resampling, then were center cropped to 224 $\times$ 224 pixels. 

For the proposed IbM2, we always sampled 200 virtual examples for every original training instance (\ie, $R=200$). We initially set the value of accuracy threshold $T$ in Algorithm 1 of the main paper as 0.9. By preliminarily training a linear classifier on the original training set $D$ before performing Algorithm 1, we got the empirical training accuracy upper bound (denoted as $\text{ACC}_{up}$), set $T$ as the minimum of its initially predefined value (0.9 here) and $\text{ACC}_{up}$ (where $T=\min\{0.9, \text{ACC}_{up}\}$ in this case) to ensure its validity. During the process of searching for the best value of $\epsilon$, as shown in Algorithm 1, we initialized a linear classifier once, and repeatedly trained it with learning rate 1.0 for 20 epochs until the searching range became tight enough. 

Next we trained the linear classifier for IbM2 (based on $D^{\hat \epsilon}$, the set of virtual examples) or the baseline method (based on the original training set $D$). Since the evaluation accuracy is sensitive to the selected learning rate during few-shot training as shown in the literature~\cite{hu2022pushing}. In the main paper, we compared IbM2 with the baseline by showing the best accuracy among 10 experimental setting, where the sole difference was the initial learning rate---it enumerated values in the set \{0.0001, 0.0005, 0.001, 0.005, 0.01, 0.05, 0.1, 0.5, 1.0, 5.0\}. We trained a series of 10 linear classifiers with the initial learning rate being one of the 10 values, and reported the best top-1 accuracy among them. The learning rate was scaled as \textit{init\_lr} $\times$ \textit{batch\_size / 256}. In this stage, for ViT~\cite{dosovitskiy2020image} architectures, the linear classifier was trained with batch size 256 for 100 epochs. For ResNet50~\cite{he2016deep}, the linear classifier was trained with batch size 512 for 60 epochs.

\section{Best vs. Default Learning Rate}

But in practice we are not allowed use the test set to find the best initial learning rate. Furthermore, in few-shot learning, even cross validation is not possible and we cannot determine the best initial learning rate using cross validation. Hence, we first determine the default initial learning rate for both methods based on the above experiments, which is 0.005 for the baseline and 1.0 for IbM2. Then, we also compare their results using the default value as initial learning rate.

\begin{table}
   \centering
   \small
   \setcounter{table}{6} 
   \renewcommand{\arraystretch}{1.04}
        \begin{tabular}{ccccc}
        \hline
        \multirow{2}{*}{\begin{tabular}[c]{@{}c@{}}Pretraining \\ Method\end{tabular}} & \multirow{2}{*}{Backbone} & \multirow{2}{*}{IbM2} & \multirow{2}{*}{Best LR} & \multirow{2}{*}{Default LR} \\
                                                                                       &                               &                       &                              &                                    \\ \hline
        \multicolumn{1}{c|}{\multirow{4}{*}{DINO~\cite{caron2021emerging}}}                                     & \multirow{2}{*}{ViT-S/8}      &                       & 70.1                         & 68.9                               \\
        \multicolumn{1}{c|}{}                                                          &                               & $\checkmark$          & 70.6                         & 70.5                               \\ \cline{2-5} 
        \multicolumn{1}{c|}{}                                                          & \multirow{2}{*}{ViT-S/16}     &                       & 64.7                         & 63.2                               \\
        \multicolumn{1}{c|}{}                                                          &                               & $\checkmark$          & 65.2                         & 65.1                               \\ \hline
        \multicolumn{1}{c|}{\multirow{2}{*}{MoCov3~\cite{chen2021empirical}}}                                   & \multirow{2}{*}{ViT-S/16}     &                       & 58.1                         & 58.1                               \\
        \multicolumn{1}{c|}{}                                                          &                               & $\checkmark$          & 59.1                         & 58.9                               \\ \hline
        \multicolumn{1}{c|}{\multirow{6}{*}{MSN~\cite{assran2022masked}}}                                      & \multirow{2}{*}{ViT-S/16}     &                       & 67.3                         & 66.5                               \\
        \multicolumn{1}{c|}{}                                                          &                               & $\checkmark$          & 68.0                         & 68.0                               \\ \cline{2-5} 
        \multicolumn{1}{c|}{}                                                          & \multirow{2}{*}{ViT-B/4}      &                       & 75.4                         & 75.4                               \\
        \multicolumn{1}{c|}{}                                                          &                               & $\checkmark$          & 76.1                         & 76.1                               \\ \cline{2-5} 
        \multicolumn{1}{c|}{}                                                          & \multirow{2}{*}{ViT-L/7}      &                       & 74.8                         & 74.8                               \\
        \multicolumn{1}{c|}{}                                                          &                               & $\checkmark$          & 75.6                         & 75.6                               \\ \hline
        \multicolumn{1}{c|}{\multirow{2}{*}{SimCLR~\cite{chen2020simple}}}                                   & \multirow{2}{*}{ResNet50}     &                       & 50.5                         & 50.2                               \\
        \multicolumn{1}{c|}{}                                                          &                               & $\checkmark$          & 51.3                         & 50.8                               \\ \hline
        \multicolumn{1}{c|}{\multirow{2}{*}{BYOL~\cite{grill2020bootstrap}}}                                     & \multirow{2}{*}{ResNet50}     &                       & 55.7                         & 55.0                               \\
        \multicolumn{1}{c|}{}                                                          &                               & $\checkmark$          & 56.9                         & 56.6                               \\ \hline
        \end{tabular}
   \caption{Best vs. Default learning rate on 1\% ImageNet-1K semi-supervised learning. The column `Best LR' means the best top-1 accuracy (\%) by using the test set to pick up the best initial learning rate. This result is a replicate of Table 2 in the main paper. The column `Default LR' reports results using the default learning rate for both the baseline (0.005) and IbM2 (1.0). It is clear that in all cases, IbM2 outperforms the baseline. More importantly, even when the baseline uses the best initial learning rate chosen on the test set, while IbM2 only uses the default learning rate, IbM2 consistently outperforms the baseline in all cases.}
   \label{table:tab7_default}
\end{table}

As shown in Table~\ref{table:tab7_default}, IbM2 outperforms the baseline in \emph{all} cases. More importantly, even when the baseline uses the best initial learning rate (chosen on the test set), while IbM2 only uses the default learning rate (without using the test set), IbM2 consistently outperforms the baseline in all cases, too.

Hence, IbM2 has a clear and significant advantage over the baseline method. 

\section{Setup for Traditional FSL}

In this setup, the pretrained models were generated by different approaches~\cite{hu2022pushing,chen2021meta,mangla2020charting} on the base set. For PMF~\cite{hu2022pushing}, following their guidance, we resized the input images into 224 $\times$ 224 pixels while keeping them in small resolutions for other two baselines~\cite{chen2021meta,mangla2020charting} (80 $\times$ 80 pixels in \textit{mini}-ImageNet~\cite{vinyals2016matching} and 32 $\times$ 32 pixels in CIFAR-FS~\cite{bertinetto2018meta}).

The initial value of $T$ in the proposed IbM2 was set as 0.999. During the search for $\epsilon$, the linear classifier was initialized once and trained for 50 epochs with learning rate 1.0 at each search step. To get the final linear classifier, we trained a new linear classifier with 200 epochs for a 5-way few-shot task on training set of baseline (original training set) and IbM2 (virtually sampled training set after the optimal $\hat \epsilon$ was got), respectively. 

As suggested in PMF~\cite{hu2022pushing}, we sampled a small number (20) of extra few-shot episodes sharing very similar semantics with the evaluated ones from the novel split on each scenario, in order to select the learning rate ranged in \{0.00001, 0.0001, 0.001, 0.01, 0.1, 1\} for the final training stage of IbM2 or the baseline. The selected optimal learning rate was used to train all 500 episodes of that scenario.

For PMF~\cite{hu2022pushing}, we discarded its original fine-tuning pipeline of updating weights of the  whole backbone. We only learned the linear classifier on top of the frozen features as in pFSL setting to make the comparison fairer.

We performed experiments on PMF~\cite{hu2022pushing} using ViT~\cite{dosovitskiy2020image} and ResNet50~\cite{he2016deep}, $S2M2_R$~\cite{mangla2020charting} using WRN-28-10~\cite{wrn_28_10} and Meta-Baseline~\cite{chen2021meta} using ResNet12~\cite{he2016deep} as the backbone. The other implementation details of traditional FSL are the same with pFSL as aforementioned.

\end{document}